\documentclass[runningheads]{llncs}

\usepackage{graphicx}
\usepackage{graphicx}
\usepackage{bm}
\usepackage{multirow}
\usepackage{tabulary}
\usepackage{array}
\usepackage{booktabs}
\usepackage{adjustbox}
\usepackage{courier}
\usepackage{color}

\usepackage[hidelinks, colorlinks = true, linkcolor = blue, urlcolor  = blue, citecolor = blue, anchorcolor = blue]{hyperref}

\begin{document}
\title{Improving Few-shot and Zero-shot \\Entity Linking with Coarse-to-Fine \\ Lexicon-based Retriever}

\titlerunning{Coarse-to-Fine Lexicon-based Retriever for Few-shot and Zero-shot EL}

\author{
	Shijue Huang\inst{1,2} \and
	Bingbing Wang\inst{1,2} \and
	Libo Qin\inst{3} \and 
	Qin Zhao\inst{1,2} \and \\
	Ruifeng Xu\inst{1,2}\thanks{Corresponding author}
}

\authorrunning{S. Huang et al.}

\institute{
	Harbin Institute of Technology (Shenzhen), Shenzhen 518000, China \and
	Guangdong Provincial Key Laboratory of Novel Security Intelligence \\Technologies, Shenzhen 518000, China \and
	School of Computer Science and Engineering, Central South University, China\\
	\email{\{22S051040,bingbing.wang\}@stu.hit.edu.cn}\\
	\email{lbqin@csu.edu.cn}, \email{\{zhaoqin,xuruifeng\}@hit.edu.cn}
}

\maketitle             
\begin{abstract}
Few-shot and zero-shot entity linking focus on the tail and emerging entities, which are more challenging but closer to real-world scenarios. The mainstream method is the ``retrieve and rerank'' two-stage framework. 
In this paper, we propose a coarse-to-fine lexicon-based retriever to retrieve entity candidates in an effective manner, which operates in two layers. 
The first layer retrieves coarse-grained candidates by leveraging entity names, while the second layer narrows down the search to fine-grained candidates within the coarse-grained ones. 
In addition, this second layer utilizes entity descriptions to effectively disambiguate tail or new entities that share names with existing popular entities.
Experimental results indicate that our approach can obtain superior performance without requiring extensive finetuning in the retrieval stage. Notably, our approach ranks the 1st in NLPCC 2023 Shared Task 6 on Chinese Few-shot and Zero-shot Entity Linking.

\keywords{Entity linking \and Few-shot and zero-shot learing}
\end{abstract}
\section{Introcuction}
 Entity linking is a crucial task in natural language processing that involves associating an ungrounded mention in text with its corresponding entity in a knowledge base, thereby facilitating language understanding. 
  Entity linking serves as a fundamental component for various downstream applications, including question answering~\cite{fevry-etal-2020-entities}, knowledge base completion~\cite{6823700,article}, text generation~\cite{puduppully-etal-2019-data} and end-to-end task-oriented dialogue system~\cite{qin-etal-2020-dynamic}.
However, existing entity linking systems face challenges when dealing with newly created or tail entities that share names with popular entities. To address this challenge, the few-shot and zero-shot entity linking task has been proposed, aiming to enhance models' ability to accurately link against the less popular and emerging entities~\cite{logeswaran-etal-2019-zero}.

Recent research on few-shot and zero-shot entity linking task has primarily employed a ``retrieve and rerank" two-stage framework, owing to the vast number of entities present in knowledge bases. In this framework, the first stage involves selecting candidate entities for a given mention, while the second stage reranks these candidates and selects the most probable entity.
Specifically, Logeswaran et al.~\cite{logeswaran-etal-2019-zero} explore deep cross-attention within candidate rerank stage. Wu et al.~\cite{wu-etal-2020-scalable} adopt a BERT architecture based two-stage approach for the zero-shot linking. And to leverage additional information from entity embeddings, Xu et al.~\cite{10.1007/978-3-031-17189-5_19} consider entities as input tokens and introduce a LUKE-based cross-encoder in the reranking stage. Moreover, Xu et al.~\cite{10.1145/3539597.3570418} enhance the dual encoder model in the retrieve stage by incorporating the Wikidata type system as an auxiliary supervision task. And they also release a new benchmark in Chinese.

In this paper, we propose a coarse-to-fine lexicon-based retriever that improves the few-shot and zero-shot entity linking in an effective manner.
Specifically, our proposed approach consists of two layers of lexicon-based retriever. In the first layer, two separate BM25 models~\cite{INR-019}, namely the \texttt{AT-BM25} and \texttt{KB-BM25}, are employed to retrieve coarse-grained candidate entities based on the entity names from the alias table and the knowledge base, respectively. Subsequently, in the second layer, another BM25 model called \texttt{Description-BM25} is introduced to retrieve fine-grained candidates from the previously obtained coarse-grained candidates by leveraging the entity descriptions. This step utilizes the entity description as detailed context to disambiguate tail entities that share names with existing popular entities.
To further refine the retrieved candidates, we employ a BERT-based dual-encoder for reranking. Finally, we propose an ensemble method to aggregate the results of the reranker with the outputs from the three BM25 models in the coarse-to-fine lexicon-based retriever. This ensemble approach ensures more robust predictions.

Experimental results demonstrate significant improvement of our coarse-to-fine lexicon-based retriever, even without extensive finetuning in retrieve stage. Notably, our approach achieves the 1st place in NLPCC 2023 Shared Task 6.

 The main contributions of our work are summarized as follows:
 
  \begin{itemize}
 	 	\item We propose a coarse-to-fine lexicon-based retriever based on BM25 to improve few-shot and zero-shot entity linking, which offers a solution to mitigate the computational burden while improving retrieve accuracy.
 	\item We introduce an ensemble method to aggregate the results from both retrieve and rerank stage, resulting in more robust predictions.
  \item Empirical results and analyses indicate the effectiveness of our approach in few-shot and zero-shot entity linking. And our approach ranks the 1st in NLPCC 2023 Shared Task 6.
 \end{itemize}
\section{Related Work}
\subsubsection{Entity Linking.}
Entity linking (EL) refers to the task of associating mentions with entries in a designated database or dictionary of entities. In order to achieve accurate and efficient identification of target entities from extensive knowledge bases (KBs), most entity linking systems adopt a two-stage approach known as ``retrieve and rerank." This contains a retriever that retrieves candidate entities from the knowledge base, followed by a reranker that reorders the candidates and selects the most probable entity.
In traditional entity retrieval, existing methods typically rely on simple techniques such as frequency information~\cite{yamada-etal-2016-joint} or sparse-based models~\cite{INR-019} to quickly retrieve a small set of candidate entities. For the ranking stage, neural networks have been widely utilized to calculate the relevance score between mentions and entities~\cite{10.1145/3308558.3313517,kolitsas-etal-2018-end}.

Recently, the emergence of pre-trained language models (PLMs) has led to their extensive adoption in both retrieve and rerank stages of entity linking. For instance, 
Logeswaran et al.~\cite{logeswaran-etal-2019-zero} employ a cross-encoder architecture and introduce deep cross-attention in the candidate ranking stage, demonstrating significance of attention between the mention-context pairs and entity descriptions.
Wu et al.~\cite{wu-etal-2020-scalable} explore a BERT-based two-stage approach for zero-shot linking.
Xu et al.~\cite{10.1007/978-3-031-17189-5_19} leverage additional information encoded in entity embeddings by considering entities as input tokens in the rerank stage, proposing an entity-enhanced cross-encoder based on LUKE.
Wiatrak et al.~\cite{wiatrak-etal-2022-proxy} propose a proxy-based metric learning loss coupled with an adversarial regularizer, providing an efficient alternative to conduct hard negative sampling in candidate retrieve stage.
In comparison to these existing works, our approach do not need large-scale finetuning for candidate retrieval, effectively reducing computational overhead while maintaining effectiveness.

\subsubsection{Few-shot and Zero-shot Entity Linking. }
Entity linking on temporally evolving knowledge bases (KBs) presents a formidable challenge in zero-shot settings. To address this, Hoffart et al.\cite{10.1145/2566486.2568003} introduce a method for entity linking on emerging entities, particularly dealing with the ambiguity surrounding their names. More recently, Logeswaran et al.\cite{logeswaran-etal-2019-zero} propose the zero-shot entity linking task to evaluate the generalization capability of entity linking systems, making minimal assumptions. They also release an English zero-shot entity linking dataset. In the context of Chinese entity linking, Xu et al.~\cite{10.1145/3539597.3570418} introduce a challenging multi-domain benchmark that utilizes Wikidata as the KB. 
In contrast to  most of the existing works, our research focuses specifically on few-shot and zero-shot entity linking in the Chinese language.

\section{Methodology}
In this section, we introduce our approach for few-shot and zero-shot entity linking. 
Given a knowledge base $E$ consisting of entities, each characterized by a name $e_i$ and a description $d_i$. The objective of this task is to determine the most suitable entity for a given mention $m_i$ within a document $doc_i$.

Our approach follows the ``retrieve and rerank'' two-stage framework. In the retrieve stage, we propose a coarse-to-fine lexicon-based retriever that efficiently retrieves candidate entities without the need for time-consuming large-scale finetuning ($\S \ref{sec:retriver}$). In the rerank stage, we adopt a BERT-based dual-encoder to reevaluate the retrieved candidate entities and select the most appropriate entity ($\S \ref{sec:rerank}$). Finally, we introduce an ensemble method to combine the results obtained from both stages, thereby producing more robust predictions ($\S \ref{sec:ensemble}$).

\subsection{Coarse-to-Fine Lexicon-based Retriever} \label{sec:retriver}

\begin{figure}[t]
	\centering
	\includegraphics[width=0.95\textwidth]{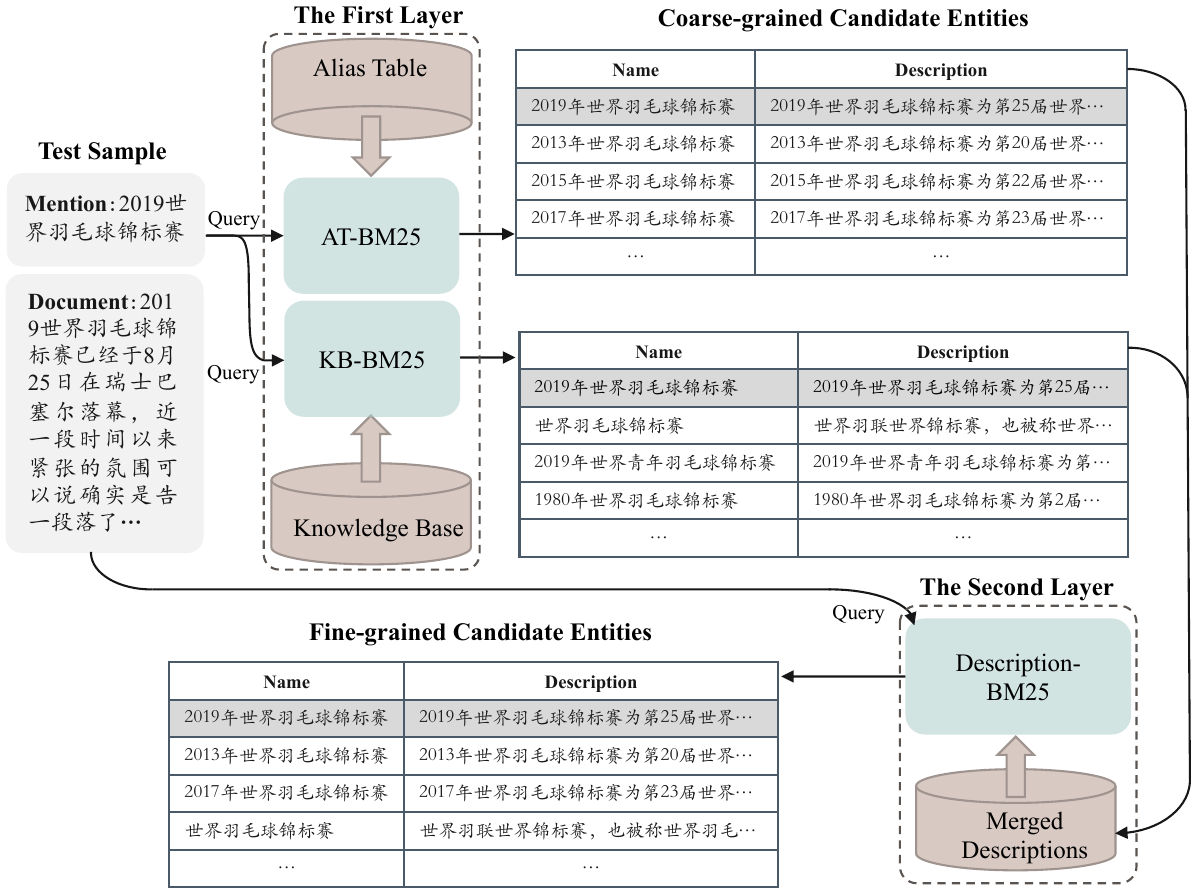}
	\caption{The architecture of coarse-to-fine lexicon-based retriever, where the first layer contains \texttt{AT-BM25} and \texttt{AT-BM25} to retrieve coarse-grained candidate entities from alias table and knowledge base. And the second layer utilizes the \texttt{Description-BM25} to obtain fine-grained candidate entities from coarse-grained ones.} \label{fig1}
\end{figure}

The overall architecture of our proposed coarse-to-fine lexicon-based retriever is illustrated in \figurename~\ref{fig1}. It contains two layers to retrieve candidate entities in a coarse-to-fine manner.
\subsubsection{The First Layer (Coarse Stage).}
Given a text sample $(m_i, doc_i)$, an entity set $E = \{e_i, d_i\}_{i=1}^N$, and an alias table $AT=\{m_j, e_j\}_{j=1}^{M}$ that defines the probability of a mention $m_j$ linking to an entity $e_j$
, we construct the retriever in two layers.
In the first layer, we employ two separate BM25 models, namely \texttt{AT-BM25} and \texttt{KB-BM25}. The \texttt{AT-BM25} model uses the entity names in the alias table as the corpus, while the \texttt{KB-BM25} model utilizes that in the knowledge base. By treating the mention $m_i$ in the test sample as a query, we utilize these two models to retrieve coarse-grained candidate entities as follows:
\begin{eqnarray}
	D_{AT} &=& \{m_j, m_j \in AT\}, \\
	D_{KB} &=& \{e_i, e_i \in E\}, \\
	Cand_{AT} &=& \texttt{AT-BM25}(m_i, D_{AT}), \\
	Cand_{KB} &=& \texttt{KB-BM25}(m_i, D_{KB}),
\end{eqnarray}
where $D_{AT}$ and $D_{KB}$ are the corpus to build \texttt{AT-BM25} and \texttt{KB-BM25}, respectively;  $Cand_{AT}$ and $Cand_{KB}$ are the retrieved coarse-grained candidate entities from alias table and knowledge base.

\subsubsection{The Second Layer (Fine Stage).}
To further disambiguate the tail entities based on the detailed mention context, we treat the document $doc_i$ of the test sample as query and merge the obtained coarse-grained candidate entities $Cand_{AT}$ and $Cand_{KB}$ into a non-repeated set $Cand_1$. We then introduce the second layer BM25 model, called \texttt{Description-BM25}, which utilizes the descriptions from the obtained coarse-grained candidate entities as the corpus. This step retrieves the fine-grained candidate entities as follows:
\begin{eqnarray}
Cand_{1} &=&  Cand_{AT} \cup Cand_{KB}, \\
D_{des} &=&  \{d_i, d_i \in Cand_{1}\}, \\
Cand_2 &=& \texttt{Description-BM25}(doc_i, D_{des}), 
\end{eqnarray}
where $Cand_{1}$ are the merged coarse-grained candidate entities; $D_{des}$ is the union of descriptions from $Cand_{1}$, which is used to build  \texttt{Description-BM25}; $Cand_2$ is the obtained fine-grained candidate entities.

\subsection{BERT-based Dual Encoder} \label{sec:rerank}
\begin{figure}[t]
	\centering
	\includegraphics[width=0.95\textwidth]{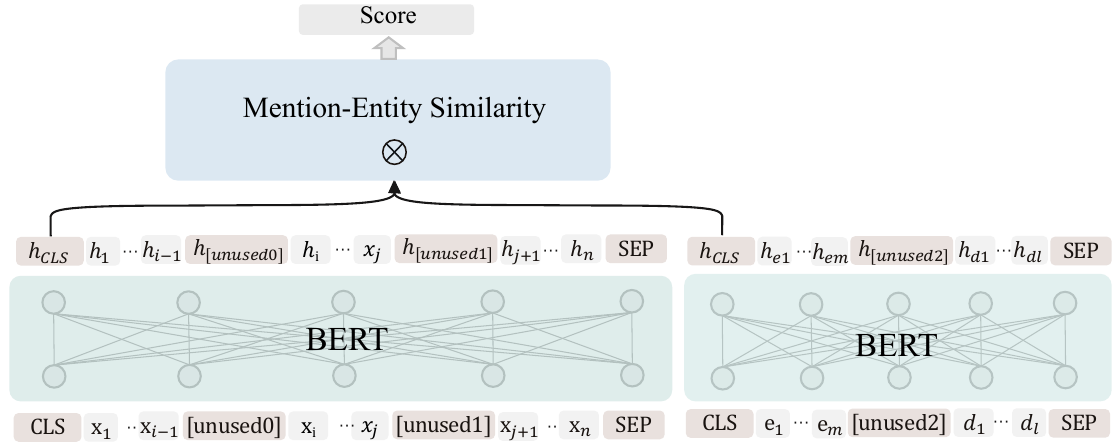}
	\caption{The architecture of BERT-based dual-encoder. Two BERT encoders are employed to encode the mention within context and the entity-description pair separately. And the mention-entity similarity is computed by dot product.} \label{fig2}
\end{figure}

To rerank the obtained coarse-grained candidate entities and get the most proper entity, we follow previous works~\cite{botha-etal-2020-entity,wu-etal-2020-scalable} to train a BERT-based dual encoder. 
This approach offers scalability benefits, as the entity embeddings can be pre-computed and stored, allowing for fast retrieval and similarity score computation using dot product.
\figurename~\ref{fig2} illustrates the architecture of the BERT-based dual encoder. 
Given a document $\{x_1, x_2, ... x_n\}$ with $n$ tokens, where a mention $m=\{x_i, x_{i+1}, ... x_j\}$ is present, the input sequence for the mention is constructed as follows:
\begin{eqnarray}
	X_1 = \texttt{[CLS]} x_1, ..., x_{i-1} \texttt{[unused0]} x_i,... ,x_j \texttt{[unused1]} x_{j+1},...,x_n \texttt{[SEP]},
\end{eqnarray}
where \texttt{[unused0]} and \texttt{[unused1]} are are special tokens stand of mention start and end, respectively. 
Following Humeau et al.~\cite{Humeau2020Poly-encoders:}, we feed the input sequence of the mention into the BERT encoder~\cite{devlin-etal-2019-bert} to obtain the representation of the \texttt{[CLS]} token in the last layer:
\begin{eqnarray}
	\bm y_m = \texttt{BERT}(X_1),
\end{eqnarray}
where $\bm y_m$ is the representation of given mention.

For a given entity $e$, which includes the entity name $\{e_1, ..., e_{m}\}$ and description $\{d_1, ... , d_l\}$, the input sequence is created by concatenating the entity name and description:
\begin{eqnarray}
	X_2 = \texttt{[CLS] $e_1,...,e_{m}$ [unused2] $d_1,...,d_l$ [SEP]},
\end{eqnarray}
where \texttt{[unused2]} is a special token to separate the entity name and description. 
Similar to the mention encoding approach, we feed the input sequence of entity into the BERT encoder to obtain the representation of the entity:
\begin{eqnarray}
	\bm y_e = \texttt{BERT}(X_2),
\end{eqnarray}
where $\bm y_e$ is the representation of entity.

Finally, the score of the mention and entity pair is calculated as the dot product of their representations:

\begin{eqnarray}
	s(m, e) = \bm y_m \cdot \bm y_e
\end{eqnarray}

\subsubsection{Optimization.} We apply the standard cross-entropy loss to train the BERT-based dual encoder:
\begin{eqnarray}
	\mathcal{L} = -\frac{1}{N} \sum_{n=1}^{N} y_n \cdot log \hat{y_n},
\end{eqnarray}
where $N$ is the number of total samples, $y_n$ is the golden label, and $\hat{y_n}$ is the predicted distribution.

\subsubsection{Inference.} 
During the inference phase, we utilize the BERT-based dual encoder to rerank the combination of coarse-grained and fine-grained candidate entities ($Cand_1 \cup Cand_2$) in the retrieval stage to avoid missing retrievals and error propagation. Moreover, to enhance computational efficiency, we pre-compute $\bm{y_e}$ for each $e \in E$ and store all the entity embeddings.

\subsection{Ensemble Method} \label{sec:ensemble}
Due to we focus on entity linking in few-shot and zero-shot scenarios, 
where a majority of the entity mentions are either unseen or only a few of them are encountered during training. This could introduce a bias in the rerank stage, because there is a huge gap between the training set and the few-shot and zero-shot test set. To mitigate this bias, we propose an ensemble method that leverages information from both the retrieve and rerank stages, aiming to enhance the robustness of predictions.

To achieve this, we employ an ensemble strategy that combines four results obtained from the retrieve and rerank stages, include:
\begin{itemize}
	\item The top-ranked result from the \texttt{AT-BM25} retrieval process;
	\item The top-ranked result from the \texttt{KB-BM25} retrieval process;
	\item The top-ranked result from the \texttt{Description-BM25} retrieval process;
	\item The top-ranked prediction from the BERT-based dual encoder;
\end{itemize}

In our ensemble strategy, the final predicted result is determined by considering above four predicted results. If two or more results are identical, we select the same result as the final prediction. In cases where the four predicted results differ, we choose the output from the BERT-based dual encoder as the final prediction. Additionally, there may has a unique 2:2 situation arising during voting. In this case, we opt for the prediction that incorporates the result derived from the BERT-based dual encoder.

\section{Experiments}

\subsection{Data}

\subsubsection{Hansel Dataset.}
Hansel~\cite{10.1145/3539597.3570418} is a benchmark dataset for few-shot and zero-shot entity linking in simplified Chinese. The training set of Hansel is derived from Wikipedia and the test set consists of Few-Shot (FS) and Zero-Shot (ZS) settings. The FS setting focuses on tail entity linking, while the ZS setting aims to evaluate the zero-shot generalization to emerging and tail entities. The dataset statistics are presented in Table~\ref{tab:stats_hansel}.

\subsubsection{Knowledge Base.}
In order to reflect the realistic scenario of knowledge bases evolving over time, the Hansel dataset divides Wikidata entities into two sets, namely the Known and New sets, based on two historical dumps:
\begin{itemize}
	\item Known Entities ($E_{known}$) are Wikidata entities from the dump on August 13, 2018. And our models are trained using $E_{known}$ as the knowledge base.
	\item New Entities ($E_{new}$) refer to Wikidata entities from the dump on March 15, 2021, which are not present in $E_{known}$. These entities are added to Wikidata between 2018 and 2021 and unseen during training on the 2018 data, representing a zero-shot setting.
\end{itemize}

\subsubsection{Alias Table.}
 For both $E_{known}$ and $E_{new}$, Hansel constructs an alias table by extracting information from Wikipedia dated March 1, 2021. And this table is generated by parsing Wikipedia's internal links, redirections, and page titles.

\begin{table}[t]
	\centering
		\caption{
		Statistics of the Hansel dataset.
		\label{tab:stats_hansel}}
	\begin{tabular}{l ccc ccc ccc} \toprule
		& \multicolumn{3}{c}{\bf \# Mentions} 
		& \multicolumn{3}{c}{\bf \# Documents} 
		& \multicolumn{3}{c}{\bf \# Entities} 
		\\ 
		\cmidrule(lr){2-4}  \cmidrule(lr){5-7}  \cmidrule(lr){8-10}  
		& In-KB & NIL & Total & In-KB & NIL & Total & $E_{known}$ & $E_{new}$   & Total \\ 
		\midrule
		Train & 9.89M & - & 9.89M & 1.05M & - & 1.05M  & 541K & -     & 541K \\
		Validation & 9,677 & - & 9,677 & 1,000 & - & 1,000  & 6,323 & -     & 6,323 \\
		\bottomrule
	\end{tabular}
%}
\end{table}

\subsection{Experimental Settings}
In the retrieve stage, we use rank-BM25\footnote{https://github.com/dorianbrown/rank\_bm25} implementation of BM25 algorithm for the coarse-to-fine lexicon-based retriever. And the number of retrieved candidate entities for \texttt{AT-BM25}, \texttt{KB-BM25} and \texttt{Description-BM25} are all set 10.

In the rerank stage, our model is implemented using the Huggingface Library~\cite{wolf2020huggingfaces}, with \texttt{bert-base-chinese}
 as backbone model for the BERT-based dual encoder. 
We use AdamW~\cite{loshchilov2018decoupled} to optimize the parameters of model. Due to the large amount of training data, the training epoch is set to 1 and the batch size is set 384. The learning rate is set $5e^{-5}$. And the max sequence length of mention and entity are both set 128. All experiments are conducted on Tesla V100 GPUs.

\subsection{Main Results}
The evaluation metric employed in is accuracy, and the evaluation results on test set are provided by the organizers. 
The main results in shown in \tablename~\ref{tab:main} , our approach achieves 1st place in  NLPCC 2023 Shared Task 6 on Chinese Few-shot and Zero-shot Entity Linking.

\begin{table}[t]
	\centering
	\caption{
		The results for NLPCC 2023 Shared Task 6.
		\label{tab:main}}
	\setlength{\tabcolsep}{7mm}{
	\begin{tabular}{l|c} 
		\toprule
	System name  & Accuracy    \\
	\midrule
	$\bm{Ours}$ & 0.6915 \\
	ITNLP & 0.6009 \\
	YNU-HPCC & 0.5319 \\
	\bottomrule
	\end{tabular}
}
\end{table}

It is evident from the results that our approach significantly outperforms the systems ranked second and third, with an accuracy advantage of 9.06\% and 15.96\%, respectively. This notable improvement can be attributed as follows:
\begin{itemize}
	\item  The proposed coarse-to-fine lexicon-based retriever can accurately retrieve candidate entities, which plays a crucial role in our overall approach and forms a solid foundation for subsequent processes.
	
	\item The utilization of a BERT-based dual encoder allows for effective reranking of candidate entities, thereby enhancing the final prediction result.
	
	\item Our proposed ensemble method effectively combines the information from both retrieve and rerank stages to disambiguate entities, which bridges the gap between training and few-shot or zero-shot test setting.
\end{itemize}
%And this observation validates the effectiveness of our proposed approach.

\subsection{Performance of Retrieve Stage}
In order to gain a deeper understanding of how our coarse-to-fine lexicon-based retriever improves few-shot and zero-shot entity linking, we present a comprehensive analysis of the retrieve stage's performance. Specifically, we retrieve 10 candidate entities for all BM-25 models and report the recall at the top-1, top-5, and top-10 levels on the test set.

\begin{table}[t]
	\centering
	\caption{
		The performance of three used lexicon-based retriever. The r@1, r@5 and r@10 represent the top-1 recall, top-5 recall and top-10 recall, respectively.
		\label{tab:re}}
\setlength{\tabcolsep}{7mm}{
	\begin{tabular}{l|ccc} 
		\toprule
		Retriever  & r@1
& r@5 & r@10 \\
		\midrule
		AT-BM25 & 0.5232 & 0.8245 &0.8626 \\
		KB-BM25 & 0.4846 & 0.6015 & 0.6614 \\
		Description-BM25 & 0.3891 & 0.6904 & 0.8341\\
		\bottomrule
	\end{tabular}
}
\end{table}

The results are illustrated in \tablename~\ref{tab:re}. We can have the following observations:
Firstly, the \texttt{AT-BM25} model achieves the best performance across all metrics, indicating that the prior knowledge contained in the alias table significantly benefits the few-shot and zero-shot entity linking. We speculate that this improvement stems from the fact that newly created or tail mentions in the test set may share similar names with certain aliases.

Secondly, the \texttt{KB-BM25} model exhibits the poorest performance. This observation aligns with our expectations since most mentions do not strictly match entity names in the knowledge base. However, the results obtained from the \texttt{KB-BM25} model still contribute because mentions link accurately when they match entity names in the knowledge base.

Lastly, the \texttt{Description-BM25} model does not exhibit significant disambiguation as expected. We attribute this to the limited maximum length of the context, which is imposed by computational resource constraints, thereby impacting the disambiguation of tail entities. Nonetheless, \texttt{Description-BM25} model provides an alternative perspective by retrieving candidate entities based on descriptions, thus also enhancing the final prediction.

\begin{table}[t]
	\centering
	\caption{
		The ablation study.
		\label{tab:ab}}
	\setlength{\tabcolsep}{7mm}{
		\begin{tabular}{l|c} 
			\toprule
			System name  & Accuracy    \\
			\midrule
			$\bm{Ours}$ & 0.6915 \\
			w/o Ensemble & 0.6791\\
			w/o AT-BM25 & 0.6713\\
			w/o KB-BM25 & 0.6791 \\
			w/o Description-BM25 & 0.6769 \\
			\bottomrule
		\end{tabular}
	}
\end{table}

\subsection{Ablation Study}
In order to validate the effectiveness of our proposed coarse-to-fine lexicon-based retriever, we conducted an ablation study by individually removing the ensemble method and the \texttt{AT-BM25}, \texttt{KB-BM25}, and \texttt{Description-BM25}.

The results are presented in \tablename~\ref{tab:ab}. We can observe that when we remove the ensemble method, the accuracy drops 1.24\%. This suggest that the ensemble method can  aggregate useful information form both retrieve and rerank stages, which results in more robust prediction.

And it can be observed that upon removing \texttt{AT-BM25}, \texttt{KB-BM25}, and \texttt{Descr\\iption-BM25}, the accuracy decreased by 2.02\%, 1.24\%, and 1.46\%, respectively. 
This observation provides evidence that all the employed BM25 models in our coarse-to-fine lexicon-based retriever contribute positively to the task of few-shot and zero-shot entity linking. Furthermore, the combination of these models leads to the retrieval of more accurate candidate entities.

\section{Conclusion}
In this paper, we present a novel approach for improving few-shot and zero-shot entity linking through a coarse-to-fine lexicon-based retriever. Our proposed method adopts the widely-used ``retrieve and rerank" framework, consisting of two stages: the coarse-to-fine lexicon-based retriever for retrieving candidate entities, and a BERT-based dual encoder for reranking the candidate entities. Moreover, we address the learning bias during the training phase of rerank stage by employing an ensemble method that combines information from both retrieve and rerank stages. The experimental results and analyses verify the effectiveness of our approach, which achieves the 1st ranking in the NLPCC 2023 Shared Task 6 on Chinese Few-shot and Zero-shot Entity Linking.

\subsubsection{Acknowledgements.}
This research was supported in part by the National Natural Science Foundation of China(62006062, 62176076), the Guangdong Provincial Key Laboratory of Novel Security Intelligence Technologies(2022B121201000\\5), Natural Science Foundation of Guangdong(2023A1515012922), and Key Technologies Research and Development Program of Shenzhen JSGG20210802154400\\001.

\bibliography{custom}
\bibliographystyle{splncs04}

\end{document}